%% file: acl_latex.tex
\pdfoutput=1

\documentclass[11pt]{article}

\usepackage[final]{acl}

\usepackage{times}
\usepackage{latexsym}

\usepackage[T1]{fontenc}

\usepackage[utf8]{inputenc}

\usepackage{microtype}

\usepackage{inconsolata}

\usepackage[utf8]{inputenc} %
\usepackage[T1]{fontenc}    %
\usepackage{hyperref}       %
\usepackage{url}            %
\usepackage{booktabs}       %
\usepackage{amsfonts}       %
\usepackage{nicefrac}       %
\usepackage{microtype}      %
\usepackage{xcolor}         %
\usepackage{graphicx}       %
\usepackage[para,online,flushleft]{threeparttable}
\usepackage{multirow}
\usepackage{amsmath}
\usepackage{array}
\usepackage{wrapfig}
\newcolumntype{C}[1]{>{\centering\let\newline\\\arraybackslash\hspace{0pt}}m{#1}}

\newcommand{\eg}{\textit{e.g., }}
\newcommand{\ie}{\textit{i.e., }}

\newcommand{\labone}{\texttt{Priority}}
\newcommand{\labtwo}{\texttt{Average}}
\newcommand{\CrossMono}{\texttt{TCM}}

\newcommand{\tten}{\texttt{en}}
\newcommand{\ttko}{\texttt{ko}}
\newcommand{\ttja}{\texttt{ja}}
\newcommand{\ttru}{\texttt{ru}}
\newcommand{\ttfr}{\texttt{fr}}
\newcommand{\ttxx}{\texttt{xx}}
\usepackage[normalem]{ulem}
\useunder{\uline}{\ul}{}
\usepackage{colortbl}

\definecolor{mygray}{gray}{0.85}
\newcommand\grayc[1]{\multicolumn{1}{>{\columncolor{mygray}}c}{#1}}
\newcommand\grayclineright[1]{\multicolumn{1}{>{\columncolor{mygray}}c|}{#1}}

\title{
Improving Multi-lingual Alignment Through Soft Contrastive Learning
}

\author{
    Minsu Park$^*$\\
    Yonsei University\\
    \texttt{0601p@yonsei.ac.kr} \\
    \And
    Seyeon Choi$^*$\\
    Yonsei University\\
    \texttt{seyeon717@yonsei.ac.kr} \\
    \And
    Chanyeol Choi \\
    Linq\\
    \texttt{jacob.choi@getlinq.com}\\
    \AND
    Jun-Seong Kim$^\dagger$ \\
    Linq\\
    \texttt{junseong.kim@getlinq.com} \\
    \And
    Jy-yong Sohn$^\dagger$ \\
    Yonsei University\\
    \texttt{jysohn1108@gmail.com}\\
}

\input{math_commands}

\begin{document}
\maketitle
\def\thefootnote{*}
\footnotetext{Equal contribution}
\def\thefootnote{$\dagger$}
\footnotetext{Corresponding authors}
\def\thefootnote{\arabic{footnote}}

\begin{abstract}
Making decent multi-lingual sentence representations is critical to achieve high performances in cross-lingual downstream tasks. In this work, we propose a novel method to align multi-lingual embeddings based on the similarity of sentences measured by a pre-trained mono-lingual embedding model. 
Given translation sentence pairs, we train a multi-lingual model in a way that the similarity between cross-lingual embeddings follows the similarity of sentences measured at the mono-lingual teacher model. Our method can be considered as contrastive learning with \textit{soft} labels defined as the similarity between sentences. Our experimental results on five languages show that our contrastive loss with soft labels far outperforms conventional contrastive loss with hard labels in various benchmarks for bitext mining tasks and STS tasks. In addition, our method outperforms existing multi-lingual embeddings including LaBSE, for Tatoeba dataset. The code is available at \url{https://github.com/YAI12xLinq-B/IMASCL}
\end{abstract}

\section{Introduction}

Learning good representations (or embeddings) of sentences and passages is crucial for developing decent models adaptive to various downstream tasks in natural language processing. %
Compared with the high quality mono-lingual sentence embeddings developed in recent years~\citep{e5-instruct,song2020mpnet}, multi-lingual sentence embeddings have a room for improvement, mostly due to the difficulty of gathering translation pair data compared to mono-lingual data. This motivated recent trials on improving the performance of multi-lingual embeddings.

One of the prominent approaches trains the multi-lingual embeddings using contrastive learning~\citep{CLinNLP, gao2021simcse}. 
Given a translation pair for different languages, this approach trains the model in a way that the embeddings for translation pairs are brought closer together, while embeddings for non-translation pairs are pushed further apart~\citep{LABSE}.
Despite several benefits of this contrastive learning approach,~\citet{ham2021semantic} pointed out that current training method ruins the mono-lingual embedding space. 
To be specific, this issue arises from the fact that existing contrastive loss treats sentences that are not exact translation pairs identically (as negative pairs), irrespective of the semantic similarity of those sentences. 

Another prominent approach is distilling mono-lingual teacher embedding space to a multi-lingual student model. The basic idea is, letting the multi-lingual embeddings of student models follow the mono-lingual embeddings of teacher model. This approach is motivated by the assumption that English embeddings are well constructed enough to guide immature multi-lingual embeddings. 
For example,~\citet{reimers2020making} proposed a distillation method using mean-squared-error (MSE) loss, which is shown to be effective in learning embeddings for low-resource languages. Another work by~\citet{LASER3} used a distillation method where the teacher is the English embedding of a multi-lingual model.
Unfortunately, existing distillation methods cannot fully utilize the translation pairs. Since conventional methods choose the most reliable English embeddings as the teacher model, translation pairs from non-English language parallel corpus are not fully leveraged.

\begin{figure}[t!] 
\centering
\includegraphics[width=\linewidth]{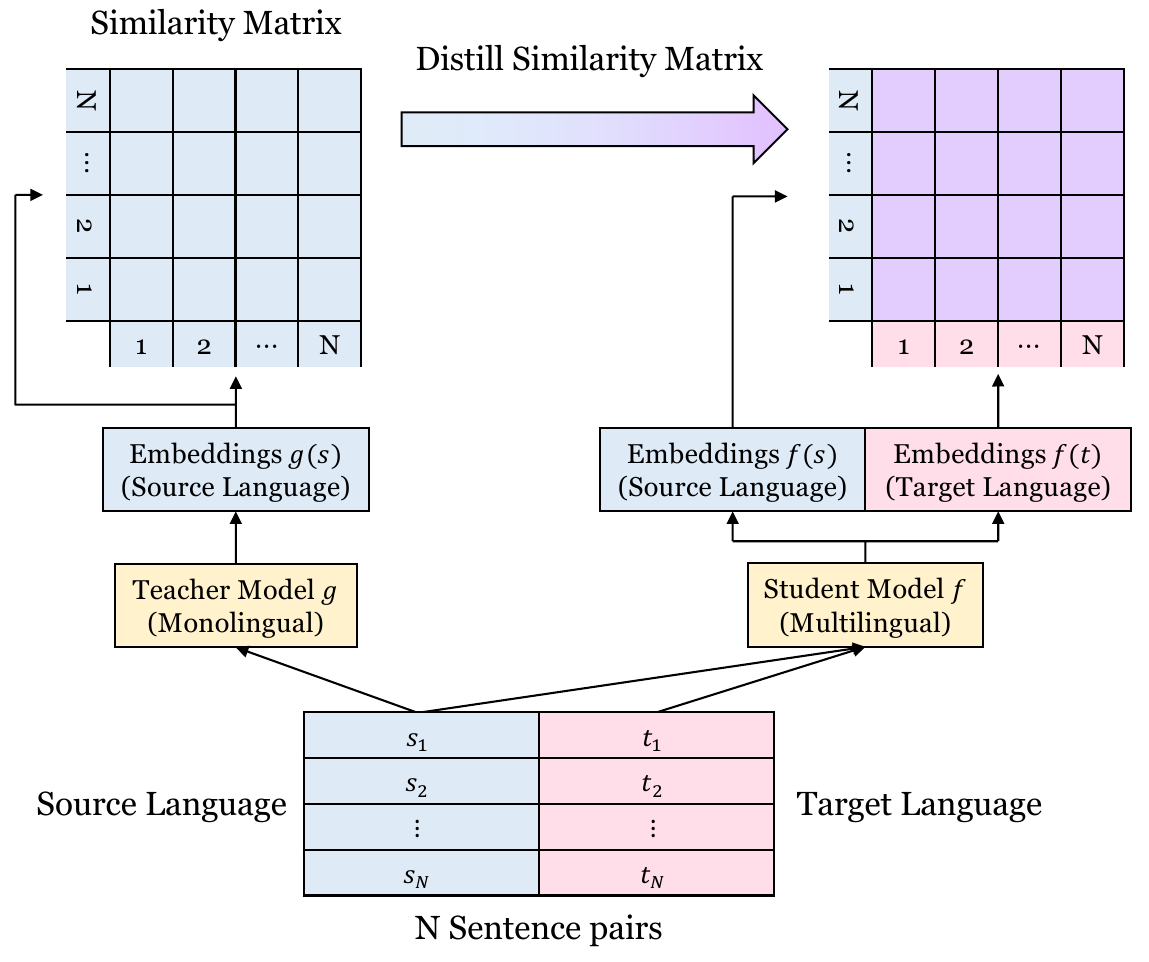}
\vspace{-2mm}
\caption{Overall framework of our method. Given $N$ sentence pairs from source/target languages, we train a multi-lingual student model $f$ by using the similarity between sentences measured by a mono-lingual teacher model $g$. Our contrastive loss function in Eq.~\ref{eq:symmetric} uses soft-label $w(i,j)$ defined in Eq.~\ref{eq:label_priority} and~\ref{eq:label_avg}. 
}
\label{fig:architecture overview}
\end{figure}

In this paper, we propose a novel distillation method for improving multi-lingual embeddings, by using \textit{soft} contrastive learning, as in Figure~\ref{fig:architecture overview}. Here, we focus on improving the sentence embeddings for symmetric tasks, when we fine-tune pre-trained encoder-only multi-lingual models by using translation pairs.  
To be specific, given $N$ translation pairs $\{(s_i, t_i)\}_{i=1}^N$, our method first computes the mono-lingual sentence similarity matrix from the teacher model. Each element of this similarity matrix is a continuous value. We distill such \textit{soft} label to the cross-lingual sentence similarity matrix computed for the multi-lingual student model. 
In other words, the anchor similarity matrix computed from the teacher model is used as a pseudo-label for contrastive loss.

Our main contributions are as follows:
\begin{itemize}
    \item We propose a novel method of fine-tuning multi-lingual embeddings by distilling the sentence similarities measured by mono-lingual teacher models. Compared with the conventional contrastive learning which uses hard labels (either positive or negative for sentence pairs), our method chooses \textit{soft} labels for measuring the sentence similarities.
    \item Compared with conventional contrastive learning and monolingual distillation method using MSE, our soft contrastive learning has much improved performance in bitext mining tasks, Tatoeba, BUCC and FLORES-200, for five different languages. 
    \item For Tatoeba, our method outperforms existing baselines including LaBSE, LASER2 and MPNet-multi-lingual.
\end{itemize}

\section{Related Works}

Constructing multi-lingual embedding has been actively studied for recent years~\citep{LASER3,LASER,2023SONAR}. 
For example, LaBSE~\citep{LABSE} shows remarkable bitext retrieval performances, which is first pretrained with masked language modeling (MLM)~\citep{devlin2018bert} and translation language modeling (TLM)~\citep{XLM} tasks, and then fine-tuned with translation pairs using contrastive loss, i.e. translation ranking task. Also, mUSE~\citep{yang2019mUSE} uses translation based bridge tasks from \citet{chidambaram2018learning} to make a multilingual embedding space. In short, mUSE~\citep{yang2019mUSE} is trained for translation ranking task with hard negatives.

\citet{reimers2020making} introduced distilling the mono-lingual embedding of a teacher model (using sBERT~\citep{sBERT}) to the multi-lingual embedding of a student model (using XLM-R~\citep{XLM-Roberta}) with MSE loss, which enables a good bitext retrieval performance only with small amounts of parallel data. 
Several follow-up papers~\citep{2023SONAR,LASER3} achieved good performances by using this distillation approach.
For example,~\citet{LASER3} successfully improved the performance on low-resource languages with the aid of the distillation approach. 
Compared with existing distillation methods, our work distills the similarities between sentences measured by mono-lingual embeddings, instead of directly distilling the mono-lingual embedding space of the teacher model.

\section{Proposed Method}\label{sec:method}

Suppose we are given $N$ translation pairs, denoted by $(s_1, t_1)$, $(s_2, t_2)$, $\cdots$, $(s_N, t_N)$, where $s_i$ is the $i$-th sentence in the source language and $t_i$ is the corresponding sentence in the target language. 
Throughout the paper, the language with bigger corpus size (used for training the teacher model) is chosen as the source language.  
We train a multi-lingual student model $f$ by using the similarities between mono-lingual sentences measured by a teacher model $g$. Here, we either distill from a mono-lingual model $g$ or distill the sentence similarities for a single language, measured by a multi-lingual model $g$.  

\paragraph{Training Cross-lingual Space Only}

Given the teacher model $g$, we train the cross-lingual space of the student model $f$ in a way that 
\begin{align}
\label{eqn:trainpurpose}
    \texttt{sim}_g(s_i, s_j) &\approx \texttt{sim}_f(s_i, t_j), \nonumber\\ 
    \texttt{sim}_g(s_i, s_j) &\approx \texttt{sim}_f(t_i, s_j)
\end{align}
\ie the similarity of $i$-th sentence and $j$-th sentence is maintained across different language pairs. Throughout the paper, we use the cosine similarity with temperature parameter $\tau$, denoted by \begin{equation*}%
    \texttt{sim}_g(s_i, s_j) = \frac{\texttt{cos}(g(s_{i}), g(s_{j}))}{\tau}.
\end{equation*}
Such objective in Eq.~\ref{eqn:trainpurpose} is reflected in our bi-directional contrastive loss:
\begin{equation} \label{eq:symmetric}
    L_{cross} = L_{row} + L_{col},
\end{equation}
following the convention used in recent works including~\citep{radford2021CLIP}, where 
\begin{equation*}
\small
    L_{row}=-\frac{1}{N}\sum_{i=1}^{N} \sum_{j=1}^N w(i, j)\log( \frac{e^{\texttt{sim}_f(s_{i}, t_{j})}}{\sum_{n=1}^N e^{\texttt{sim}_f(s_{i}, t_{n})}})
\end{equation*}
and
\begin{equation*}
    L_{col}=-\frac{1}{N}\sum_{i=1}^{N} \sum_{j=1}^N w(i,j)\log( \frac{e^{\texttt{sim}_f(s_{i}, t_{j})}}{\sum_{n=1}^N e^{\texttt{sim}_f(s_{n}, t_{j})}}).
\end{equation*}

Here, $w(i,j)$ is the label (either hard or soft) defined by the similarity between $s_i$ and $s_j$.
For example, the standard contrastive loss used in LaBSE~\citep{LABSE} and mE5~\citep{wang2024me5} has the form of Eq.~\ref{eq:symmetric} where 
\begin{align}\label{eqn:hard-label}
    w(i, j) = \begin{cases}
            1 & \text{ if } i = j \\
            0 & \text{ otherwise } \\    
    \end{cases}
\end{align}
\ie two sentences ($s_i$ and $s_j$) are considered as a positive pair only if $i=j$, and labeled as a negative pair otherwise.

Since the na\"ive method of using the \textit{hard} label (shown above) cannot fully capture the semantic relationship between different sentence pairs, we propose two novel ways of defining \textit{soft} label $w(i,j)$, which are denoted by (1) \labone{} label, and (2) \labtwo{} label. For the first option, we use the similarity of sentences measured at the source language only. In other words, \labone{} label is defined as  
\begin{equation}\label{eq:label_priority}
    w(i, j) = \frac{e^{\texttt{sim}_g(s_{i}, s_{j})}}{\sum_{n=1}^N e^{\texttt{sim}_g(s_{i}, s_{n})}},
\end{equation}
which applies the softmax function on the similarity of sentences in the source language, measured at the teacher model $g$.

Our second option considers the mono-lingual embedding spaces for both the source and the target languages. To be specific, \labtwo{} label is defined as
\begin{equation}\label{eq:label_avg}
    w(i, j) = \frac{e^{(\texttt{sim}_g(s_{i}, s_{j})+\texttt{sim}_g(t_{i}, t_{j})) / 2}}{\sum_{n=1}^N e^{(\texttt{sim}_g(s_{i}, s_{n})+\texttt{sim}_g(t_{i}, t_{n})) / 2
    }}.
\end{equation}
Note that the second labeling option only works when the teacher model is multi-lingual.

\paragraph{Training Both Cross-lingual and Mono-lingual Space (\CrossMono{})}
Recall that the loss $L_{cross}$ given in Eq.~\ref{eq:symmetric} is motivated by the objective in Eq.~\ref{eqn:trainpurpose}, which focuses on the cross-lingual sentence similarities $\texttt{sim}_f (s_i, t_j)$ and $\texttt{sim}_f (t_i, s_j)$. %
 In addition to that, we consider learning with additional mono-lingual loss
 \begin{align} \label{eq:mono}
    L_{mono}&=-\frac{1}{N}\sum_{i=1}^{N} \sum_{j=1}^N w(i,j)\log( \frac{e^{\texttt{sim}_f(s_{i}, s_{j})}}{\sum_{n=1}^N e^{\texttt{sim}_f(s_{n}, s_{j})}})\nonumber\\
    &+-\frac{1}{N}\sum_{i=1}^{N} \sum_{j=1}^N w(i,j)\log( \frac{e^{\texttt{sim}_f(t_{i}, t_{j})}}{\sum_{n=1}^N e^{\texttt{sim}_f(t_{n}, t_{j})}}),
\end{align}
the distillation loss defined by the mono-lingual similarities, $\texttt{sim}_f(s_{i}, s_{j})$ and $\texttt{sim}_f(t_{i}, t_{j})$.
This approach of using $L_{mono}$ on top of $L_{cross}$ is dubbed as Training both Cross-lingual and Monolingual space (\CrossMono{}). We use the combined loss term, denoted by 
\begin{equation} \label{eq:TMS}
    L = \lambda \cdot L_{cross} + L_{mono},
\end{equation}
where the parameter $\lambda$ controls the balance between the cross-lingual loss and the mono-lingual loss term. Note that using \CrossMono{} is orthogonal to the choice of using the \labone{} label or the \labtwo{} label.

\section{Experimental Settings}

This section describes the details of our experimental setting, for both training and evaluation.

\subsection{Training setup}
The translation pairs used for training are downloaded from OPUS\footnote{\url{https://opus.nlpl.eu}}~\citep{OPUS}, where
the volume of each language corpus is given in Appendix \ref{Appendix B.1: DATASET}. We focus on five languages: English (\texttt{en}), French (\ttfr{}), Japanese (\ttja{}), Korean (\ttko{}), and Russian (\ttru). We train two types of models, cross-lingual and multi-lingual. For each cross-lingual model, we use \texttt{en-ko}, \texttt{en-ja}, \texttt{en-ru}, and \texttt{en-fr} pairs, respectively.
For the multi-lingual model, we train with all translation pairs for five languages.

As discussed in Sec.~\ref{sec:method}, 
we consider two types of soft label $w(i,j)$: the \labone{} label in Eq.~\ref{eq:label_priority} defines the soft label by using the similarity measured at a mono-lingual embedding (for a pre-defined anchor language), while the \labtwo{} label in Eq.~\ref{eq:label_avg} uses the similarity averaged out over mono-lingual embeddings of both source and target languages. 
Note that we need to choose the anchor language (among the source and the target language), for the former one. 
By default, we set the priority of the languages based on the volume of each language corpus used in training, thus having the following order: \texttt{en}, \texttt{ru}, \texttt{ja}, \texttt{fr}, and \texttt{ko}. The anchor language is defined as the one with higher priority between language pair. We also test \CrossMono{} shown in Eq.~\ref{eq:TMS}.

Each model is trained for 30 epochs\footnote{We early stopped with Tatoeba validation set. Most of the trains were stopped at between 10 and 20 epoch.} on 2 RTX-3090 GPUs with global batch size 32. We use the cross-accelerator to expand negative samples, as described in Appendix~\ref{Appendix B.4: CA}. The initial learning rate is set to $\gamma=5\cdot 10^{-3}$, and we linearly decay the learning rate.  We use the AdamW optimizer. The 
temperature parameter is tuned on \texttt{en-ko} bilingual dataset, which results in $\tau=0.1$. The portion of cross-lingual loss in \CrossMono{} is set to $\lambda=0.1$. We apply the mixed precision training, to improve the training efficiency.

\subsection{Evaluation tasks}

\paragraph{Bitext Mining}
We evaluate our model on three bitext mining datasets, Tatoeba~\citep{LASER}, BUCC~\citep{zweigenbaum2017overview} and FLORES-200~\citep{costa2022FLORES}. Tatoeba and BUCC are English-centric translation pair benchmark datasets that are included in MTEB~\citep{MTEB}, and FLORES-200 is a $N$-way parallel benchmark dataset. Throughout the paper, we use the average accuracy measured from both directions (e.g., \tten{}→\ttko{} and \ttko{}→\tten{}) for BUCC and Tatoeba. We measure the average xSIM error rate from ~\citep{LASER3} for each languages in FLORES-200.

\paragraph{Semantic Textual Similarity (STS)}
We evaluate our model on STS datasets to examine how well mono-lingual and cross-lingual spaces are formed. We test on STS12-STS22 and the STS benchmark in MTEB~\citep{MTEB}, and measure the average spearman correlation for different languages: \tten{}, \ttko{}, \ttfr{}, \ttru{} and \tten{}-\ttfr{}. 

\section{Results}

Here we provide experimental results on bitext mining and STS tasks. First, we describe how we choose $\text{mE5}_\text{base}$~\citep{wang2024me5} architecture as the teacher and the student model, throughout our experiments. Then, we demonstrate our \textit{soft contrastive loss} outperforms existing methods, and provide the factor analysis on our method.

\subsection{Effect of the Student/Teacher Model}
\input{tables/varying-teacher}

Table~\ref{tab: varying teacher} shows the effect of the (student, teacher) model pair on the performance of our soft contrastive loss in Eq.~\ref{eq:symmetric}, instead of Eq.~\ref{eq:TMS}.
We test two student model architectures, $\text{mE5}_\text{base}$~\citep{wang2024me5} and XLM-R~\citep{XLM-Roberta}, and three teacher models, $\text{mE5}_\text{base}$~\citep{wang2024me5}, $\text{E5}_\text{base}$~\citep{2022e5}, and MPNet\footnote{\url{https://huggingface.co/sentence-transformers/all-mpnet-base-v2}}~\citep{song2020mpnet}. 
The details of how we choose such student models are described in Appendix~\ref{sec:student-model-selection}.
One can confirm that using $\text{mE5}_\text{base}$ for both teacher and student performs the best in both STS and bitext mining tasks. Thus, for the following experiments, we use $\text{mE5}_\text{base}$ for both teacher and student.

\subsection{Effectiveness of Our Method}
\input{tables/varying-loss}
\label{Effect of Loss Function}
Recall that we train a student model using the contrastive loss with \textit{soft} labels obtained from the teacher model. Thus, we denote our proposed loss as \textit{soft contrastive loss} in the upcoming tables.

Table~\ref{tab:Crosslingual-loss} shows the performances of fine-tuned models (starting from a pre-trained  $\text{mE5}_\text{base}$ student model) trained with 4 different losses, tested on 4 different language pairs (\tten{}-\ttko{}, \tten{}-\ttfr{}, \tten{}-\ttja{}, and \tten{}-\ttru{}). Note that the language pairs used for training (and testing) each model is specified in the `Lang' column of the table. 
The loss functions we used are as follows: 
\begin{itemize}
    \item our \textit{soft contrastive loss}, especially using the \labone{} label in Eq.~\ref{eq:label_priority} with \CrossMono{} in Eq.~\ref{eq:TMS}
    \item conventional contrastive loss with hard label in Eq.~\ref{eqn:hard-label}
    \item Mean Squared Error (MSE) distillation loss proposed in~\citep{reimers2020making}
    \item the loss used in mUSE~\citep{yang2019mUSE}
\end{itemize}
Note that the number of hard negatives is set to three when we use the loss used in mUSE, due to the limited GPU resources.

From Table~\ref{tab:Crosslingual-loss}, we have three major observations.
First, our \textit{soft} contrastive loss outperforms conventional \textit{hard} contrastive loss in all performance metrics in all language pairs. For example, in Tatoeba dataset, our method has up to 5.3\% accuracy gain (\eg from 86.3\% to 91.6\% for \tten{}-\ttko{} pair) compared with hard contrastive loss. 
Note that compared with the pre-trained model (shown in the gray shaded region in Table~\ref{tab:Crosslingual-loss}), additional training with hard contrastive loss sometimes harms the performance, \eg the accuracy degrades from 87.3\% to 86.3\% in Tatoeba dataset for the model trained with \tten{}-\ttko{} pair, and the STS performance degrades from 0.802 to 0.666 for the model trained with \tten{}-\ttru{} pair, which is critical. 

Second, our \textit{soft} contrastive loss provides the best performance in the bitext mining task, Tatoeba, and BUCC, for all language pairs. Compared with the pre-trained student model, additional training with soft contrastive loss improves the accuracy up to $4.3\%$.

Third, the STS performance for non-English languages is improved, after training with our soft contrastive loss. For example, after training with \tten{}-\ttfr{} translation pair, the STS performance elevates by 0.01 (from 0.781 to 0.791) when using soft contrastive loss, while the performance degrades by 0.077 (from 0.781 to 0.704) when MSE loss~\citep{reimers2020making} is used.

Table~\ref{tab:Multilingual-loss} shows the bitext mining performances of multi-lingual models, tested on five languages, \tten{}, \ttko{}, \ttja{}, \ttfr{} and \ttru{}. 
The table compares two types of models: (1) the checkpoints of popular pre-trained multi-lingual models ($\text{mE5}_\text{base}$~\citep{wang2024me5}, LASER2~\citep{LASER}, LaBSE~\citep{LABSE}, and MPNet-multilingual\footnote{\url{https://huggingface.co/sentence-transformers/paraphrase-multilingual-mpnet-base-v2}}~\citep{reimers2020making}), and (2) the $\text{mE5}_\text{base}$~\citep{wang2024me5} model fine-tuned with either our \textit{soft contrastive loss} or MSE loss. Note that for fine-tuned models, we first load pre-trained $\text{mE5}_\text{base}$ and train it with four different language pairs: \texttt{en-ko}, \texttt{en-ja}, \texttt{en-ru}, and \texttt{en-fr}).
One can confirm that our soft contrastive loss outperforms MSE loss in every bitext mining task. Moreover, our method outperforms most existing pre-trained models.

\subsection{Factor Analysis}

\paragraph{\labone{} label vs \labtwo{} label}

\input{tables/variation_crosslingual}
\input{tables/variation_multilingual}

Recall that \labone{} label and \labtwo{} label are our proposed soft-label
$w(i,j)$ in Eq.~\ref{eq:label_priority} and~\ref{eq:label_avg}, which are used to define the soft contrastive loss in Eq.~\ref{eq:symmetric}.
Tables~\ref{tab: ours variation crosslingual} and~\ref{tab:ours variation multilingual}
compare the performances of the soft-label methods.
Note that we also test \labone{}+\CrossMono{}, which is the combination of \labone{} label with 
\CrossMono{} training loss in Eq.~\ref{eq:TMS}.

Table~\ref{tab: ours variation crosslingual} shows the performance of model trained with single language pair data. 
Fine-tuning the pretrained model with either 
\labone{} label or \labtwo{} label significantly improves the performance of most bitext mining. For example, using \labone{} label improves the performance on \tten{}-\ttko{} pairs in Tatoeba dataset, from 87.3\% to 91.7\%. 
While the performances of two labeling methods (\labone{} label and \labtwo{} label) are quite similar in most cases, 
for models trained with \texttt{en-fr},
\labtwo{} label achieves STS (\ttfr{}) performance of 0.794, which is 0.019 higher than the performance of \labone{} label.
Table~\ref{tab:ours variation multilingual} shows the performance of model trained with multiple language pairs data, which shows the tendency similar to that of Table~\ref{tab: ours variation crosslingual}.

\paragraph{Effect of \CrossMono{}}
We validate the effectiveness of \CrossMono{} by comparing \labone{} and \labone{} + \CrossMono{} in Tables~\ref{tab: ours variation crosslingual} and \ref{tab:ours variation multilingual}. Table~\ref{tab: ours variation crosslingual} shows that using \CrossMono{} improves the performance significantly on STS in all language pairs. For example, when trained on \tten{}-\ttko{} pairs, adding \CrossMono{} improves the STS performances on both \tten{} (from 0.777 to 0.788) and \ttko{} (from 0.762 to 0.778).

Adding \CrossMono{} also has a positive impact in the \textit{multi-lingual} experimental setting, as shown in Table~\ref{tab:ours variation multilingual}. 
\CrossMono{} improves the performances of \labone{} labeling, for all tested bitext mining tasks. For example, the performance increases from 97.9\% to 98.3\% for BUCC dataset. 
In addition, the comparison among the variants of our method shows that \CrossMono{} is necessary to acheive the best performance, as shown in Appendix~\ref{Appendix A: full MTEB}.

\paragraph{Effect of Language Pair Selection}

Recall that Table~\ref{tab:ours variation multilingual} shows the performance of our method when trained on five languages: \tten{}, \ttfr{}, \ttko{}, \ttja{}, \ttru{}. We have two options for choosing the language pairs of the training data. The first option (denoted by `\tten{}-\ttxx{}') uses four different pairs: \texttt{en-ko}, \texttt{en-ja}, \texttt{en-ru}, and \texttt{en-fr}. The second option (denoted by `All pairs') uses all combinations, \ie ${5 \choose 2} = 10$ language pairs. Note that changing from the first option to the second option 
does not change the performance dramatically, even though it uses more than twice larger amount of translation pairs. 
Motivated by this observation, 
we conduct additional experiments to analyze the effects of language selection on the performance of our method.

Table~\ref{tab:less language-pair} compares two options for choosing the training language pair: (1) language pairs containing \tten{} or \ttfr{} (denoted by \tten{}-\ttxx{} and \ttfr{}-\ttxx{}) , and (2) language pairs containing \tten{} or \ttru{} (denoted by \tten{}-\ttxx{} and \ttru{}-\ttxx{}. Here, we use the \labone{} label in our loss term. The former option outperforms the latter option in most cases, not  only for the bitext mining tasks, but also for STS tasks. 
Interestingly, the former option outperforms the latter option for the STS task on \ttru{} language.
This provides us a sign that it is necessary to appropriately choose the language pairs for fine-tuning embeddings; further investigation on the impact of language pairs is left as a future work.\input{tables/less_languages}

\section{Conclusion}
In this paper, we proposed a method of improving multi-lingual embeddings, with the aid of the sentence similarity information measured at the mono-lingual teacher models. Our method can be considered as a variant of existing contrastive learning approach, where our method uses \textit{soft} labels defined as the sentence similarity, while existing methods use \textit{hard} labels. %
Our method shows the best performance on the Tatoeba dataset, and achieves high performances in other bitext mining tasks as well as STS tasks.

\newpage
\bibliography{custom}

\clearpage

\input{supplementary}

\end{document}

%% file: math_commands.tex
\usepackage{amsmath,amsfonts,bm}

\def\eqref#1{equation~\ref{#1}}

\def\1{\bm{1}}

\DeclareMathAlphabet{\mathsfit}{\encodingdefault}{\sfdefault}{m}{sl}
\SetMathAlphabet{\mathsfit}{bold}{\encodingdefault}{\sfdefault}{bx}{n}

%% file: tables/varying-teacher.tex
\begin{table}[t!]
\centering
\footnotesize
\setlength{\tabcolsep}{2pt}
\begin{tabular}{c|c|c|cc|cc}
Lang                         & \begin{tabular}[c]{@{}c@{}}Student\\ Model\end{tabular} & \begin{tabular}[c]{@{}c@{}}Teacher\\ Model\end{tabular} & \begin{tabular}[c]{@{}c@{}}Tatoeba\\ (\tten-\texttt{xx})\end{tabular} & \begin{tabular}[c]{@{}c@{}}BUCC\\ (\tten-\texttt{xx})\end{tabular} & \begin{tabular}[c]{@{}c@{}}STS\\ (\tten)\end{tabular} & \begin{tabular}[c]{@{}c@{}}STS\\ (\texttt{xx})\end{tabular} \\ \hline
\multirow{6}{*}{\tten-\ttko} & \multirow{3}{*}{$\text{mE5}_\text{base}$}               & $\text{mE5}_\text{base}$                                & \textbf{0.917}                                                        & -                                                                  & \textbf{0.777}                                        & \textbf{0.762}                                              \\
                             &                                                         & $\text{E5}_\text{base}$                                 & 0.907                                                                 & -                                                                  & 0.759                                                 & 0.740                                                       \\
                             &                                                         & MPNet                                                   & 0.869                                                                 & -                                                                  & 0.692                                                 & 0.685                                                       \\ \cline{2-7} 
                             & \multirow{3}{*}{XLM-R}                                  & $\text{mE5}_\text{base}$                                & 0.896                                                                 & -                                                                  & 0.704                                                 & 0.707                                                       \\
                             &                                                         & $\text{E5}_\text{base}$                                 & 0.897                                                                 & -                                                                  & 0.702                                                 & 0.702                                                       \\
                             &                                                         & MPNet                                                   & 0.864                                                                 & -                                                                  & 0.648                                                 & 0.650                                                       \\ \hline
\multirow{6}{*}{\tten-\ttfr} & \multirow{3}{*}{$\text{mE5}_\text{base}$}               & $\text{mE5}_\text{base}$                                & \textbf{0.963}                                                        & \textbf{0.982}                                                     & \textbf{0.783}                                        & 0.775                                                       \\
                             &                                                         & $\text{E5}_\text{base}$                                 & 0.956                                                                 & 0.973                                                              & 0.764                                                 & 0.782                                                       \\
                             &                                                         & MPNet                                                   & 0.944                                                                 & 0.963                                                              & 0.706                                                 & \textbf{0.785}                                                       \\ \cline{2-7} 
                             & \multirow{3}{*}{XLM-R}                                  & $\text{mE5}_\text{base}$                                & 0.951                                                                 & 0.973                                                              & 0.699                                                 & 0.744                                                       \\
                             &                                                         & $\text{E5}_\text{base}$                                 & 0.949                                                                 & 0.961                                                              & 0.692                                                 & 0.747                                                       \\
                             &                                                         & MPNet                                                   & 0.942                                                                 & 0.956                                                              & 0.637                                                 & 0.761                                                      
\end{tabular}
\vspace{-2mm}
\caption{\label{tab: varying teacher}Comparison of various combinations of student and teacher models, in terms of the bitext mining (accuracy) and STS (spearman correlation score) performances. The best performance is achieved when both teacher and student use $\text{mE5}_\text{base}$ model.}
\end{table}

%% file: tables/varying-loss.tex
\begin{table}[t!]
\centering
\scriptsize
\setlength{\tabcolsep}{1.5pt}
\begin{tabular}{c|ccccc}
Lang &
  \multicolumn{1}{c|}{Loss} &
  \begin{tabular}[c]{@{}c@{}}Tatoeba\\ (\tten-\texttt{xx})\end{tabular} &
  \multicolumn{1}{c|}{BUCC} &
  \begin{tabular}[c]{@{}c@{}}STS\\ (\tten)\end{tabular} &
  \begin{tabular}[c]{@{}c@{}}STS\\ (\texttt{xx})\end{tabular} \\ \hline
\multirow{5}{*}{\tten-\ttko} &
  \multicolumn{1}{c|}{Soft Contrastive (Ours)} &
  \textbf{0.916} &
  \multicolumn{1}{c|}{-} &
  0.788 &
  {\ul 0.778} \\
 &
  \multicolumn{1}{c|}{Hard Contrastive} &
  0.863 &
  \multicolumn{1}{c|}{-} &
  0.674 &
  0.675 \\
 &
  \multicolumn{1}{c|}{MSE~\citep{reimers2020making}} &
  {\ul 0.911} &
  \multicolumn{1}{c|}{-} &
  \textbf{0.803} &
  \textbf{0.793} \\
 &
  \multicolumn{1}{c|}{mUSE~\citep{yang2019mUSE}} &
  0.853 &
  \multicolumn{1}{c|}{-} &
  0.715 &
  0.698 \\
 &
  \grayclineright{Pretrained Model} &
  \grayc{0.873} &
  \grayclineright{-} &
  \grayc{{\ul 0.802}} &
  \grayc{0.777} \\ \hline
\multirow{5}{*}{\tten-\ttfr} &
  \multicolumn{1}{c|}{Soft Contrastive (Ours)} &
  \textbf{0.960} &
  \multicolumn{1}{c|}{\textbf{0.987}} &
  0.796 &
  \textbf{0.791} \\
 &
  \multicolumn{1}{c|}{Hard Contrastive} &
  0.937 &
  \multicolumn{1}{c|}{0.933} &
  0.675 &
  0.748 \\
 &
  \multicolumn{1}{c|}{MSE~\citep{reimers2020making}} &
  {\ul 0.959} &
  \multicolumn{1}{c|}{0.980} &
  \textbf{0.803} &
  0.704 \\
 &
  \multicolumn{1}{c|}{mUSE~\citep{yang2019mUSE}} &
  0.950 &
  \multicolumn{1}{c|}{{\ul 0.984}} &
  0.713 &
  0.771 \\
 &
  \grayclineright{Pretrained Model} &
  \grayc{0.951} &
  \grayclineright{{\ul 0.984}} &
  \grayc{{\ul 0.802}} &
  \grayc{{\ul 0.781}} \\ \hline
\multirow{5}{*}{\tten-\ttja} &
  \multicolumn{1}{c|}{Soft Contrastive (Ours)} &
  \textbf{0.956} &
  \multicolumn{1}{c|}{-} &
  0.798 &
  \textbf{-} \\
 &
  \multicolumn{1}{c|}{Hard Contrastive} &
  0.933 &
  \multicolumn{1}{c|}{-} &
  0.730 &
  - \\
 &
  \multicolumn{1}{c|}{MSE~\citep{reimers2020making}} &
  {\ul 0.949} &
  \multicolumn{1}{c|}{-} &
  \textbf{0.808} &
  - \\
 &
  \multicolumn{1}{c|}{mUSE~\citep{yang2019mUSE}} &
  0.925 &
  \multicolumn{1}{c|}{-} &
  0.742 &
  - \\
 &
  \grayclineright{Pretrained Model} &
  \grayc{0.931} &
  \grayclineright{-} &
  \grayc{{\ul 0.802}} &
  \grayc{-} \\ \hline
\multirow{5}{*}{\tten-\ttru} &
  \multicolumn{1}{c|}{Soft Contrastive (Ours)} &
  \textbf{0.951} &
  \multicolumn{1}{c|}{\textbf{0.979}} &
  0.787 &
  \textbf{0.616} \\
 &
  \multicolumn{1}{c|}{Hard Contrastive} &
  {\ul 0.949} &
  \multicolumn{1}{c|}{0.955} &
  0.666 &
  0.545 \\
 &
  \multicolumn{1}{c|}{MSE~\citep{reimers2020making}} &
  0.945 &
  \multicolumn{1}{c|}{{\ul 0.978}} &
  \textbf{0.804} &
  0.601 \\
 &
  \multicolumn{1}{c|}{mUSE~\citep{yang2019mUSE}} &
  0.944 &
  \multicolumn{1}{c|}{{\ul 0.978}} &
  0.720 &
  0.548 \\
 &
  \grayclineright{Pretrained Model} &
  \grayc{0.936} &
  \grayclineright{{\ul 0.978}} &
  \grayc{{\ul 0.802}} &
  \grayc{{\ul 0.615}}
\end{tabular}
\vspace{-2mm}
\caption{Comparison of different loss functions used for fine-tuning pre-trained student model, 
tested on bitext mining tasks and STS tasks.
The gray shaded method is the baseline which uses the pre-trained student model as it is. The best performance is indicated in bold, second most performance is indicated with an underline, throughout this paper. For Tatoeba dataset, our \emph{soft} contrastive loss outperforms all compared losses.}
\label{tab:Crosslingual-loss}
\end{table}

\begin{table}[t!]
\centering
\scriptsize
\setlength{\tabcolsep}{1.5pt}
\begin{tabular}{c|c|ccc}
Pretrained model   & Fine-tune loss & Tatoeba        & BUCC           & FLORES-200    \\ \hline
$\text{mE5}_\text{base}$                  & Soft Contrastive (Ours)       & \textbf{0.949} & {\ul 0.983} & {\ul 0.02} \\
$\text{mE5}_\text{base}$                  & MSE & 0.942          & 0.975          & 0.05          \\ \hline
\grayclineright{$\text{mE5}_\text{base}$} & \grayclineright{-}                & \grayc{0.923}  & \grayc{0.981}  & \grayc{0.16}  \\
MPNet-multilingual & -              & 0.945          & 0.970          & 0.28          \\
LASER2             & -              & 0.939          & 0.981          & 0.20          \\
LaBSE              & -              & {\ul 0.948} & \textbf{0.985} & \textbf{0.01}
\end{tabular}
\vspace{-2mm}
\caption{Comparison between existing models and fine-tuning loss on multi-lingual data, tested on bitext mining. Measured accuracy for Tatoeba and BUCC, while measuring xSIM error rate for FLORES-200.
Note that fine-tuned with MSE is the same approach as \citet{reimers2020making}. %
}
\label{tab:Multilingual-loss}
\end{table}

%% file: tables/variation_crosslingual.tex
\begin{table}[t!]
\centering
\scriptsize
\begin{tabular}{c|c|cc|cc}
Lang &
  Loss &
  \begin{tabular}[c]{@{}c@{}}Tatoeba\\ (\tten-\ttxx)\end{tabular} &
  BUCC &
  \begin{tabular}[c]{@{}c@{}}STS\\ (\tten)\end{tabular} &
  \begin{tabular}[c]{@{}c@{}}STS\\ (\ttxx)\end{tabular} \\ \hline
\multirow{4}{*}{\tten-\ttko} & \labtwo{}                           & 0.912          & -                            & 0.771                  & 0.757               \\
                             & \labone{}                          & \textbf{0.917} & -                            & 0.777                  & 0.762               \\
                             & \labone{} + \CrossMono{}                  & {\ul 0.916}    & -                            & {\ul 0.788}            & \textbf{0.778}      \\
                             & \grayclineright{Pretrained Model} & \grayc{0.873}  & \grayclineright{-}           & \grayc{\textbf{0.802}} & \grayc{{\ul 0.777}} \\ \hline
\multirow{4}{*}{\tten-\ttfr} & \labtwo{}                           & 0.959          & 0.981                        & 0.767                  & \textbf{0.794}      \\
                             & \labone{}                          & \textbf{0.963} & 0.982                        & 0.783                  & 0.775               \\
                             & \labone{} + \CrossMono{}                  & {\ul 0.960}    & \textbf{0.987}               & {\ul 0.796}            & {\ul 0.791}         \\
                             & \grayclineright{Pretrained Model} & \grayc{0.951}  & \grayclineright{{\ul 0.984}} & \grayc{\textbf{0.802}} & \grayc{0.781}       \\ \hline
\multirow{4}{*}{\tten-\ttja} & \labtwo{}                           & {\ul 0.957}    & -                            & 0.787                  & -                   \\
                             & \labone{}                          & \textbf{0.960} & -                            & 0.787                  & \textbf{-}          \\
                             & \labone{} + \CrossMono{}                  & 0.956          & -                            & {\ul 0.798}            & -                   \\
                             & \grayclineright{Pretrained Model} & \grayc{0.931}  & \grayclineright{-}           & \grayc{\textbf{0.802}} & \grayc{-}           \\ \hline
\multirow{4}{*}{\tten-\ttru} & \labtwo{}                           & \textbf{0.955} & \textbf{0.980}               & 0.775                  & 0.612               \\
                             & \labone{}                          & {\ul 0.953}    & {\ul 0.979}                  & 0.781                  & 0.607               \\
                             & \labone{} + \CrossMono{}                  & 0.951          & {\ul 0.979}                  & {\ul 0.787}            & \textbf{0.616}      \\
                             & \grayclineright{Pretrained Model} & \grayc{0.936}  & \grayclineright{0.978}       & \grayc{\textbf{0.802}} & \grayc{{\ul 0.615}}
\end{tabular}
\vspace{-2mm}
\caption{Comparison of each variation trained with cross-lingual data, in terms of the bitext mining 
and STS performances.
\labone{} shows slightly better performance than \labtwo{} in bitext mining tasks, except for \texttt{en-ru}. Applying \CrossMono{} enhances the STS performance, better than a pre-trained model for the non-English language.
}
\label{tab: ours variation crosslingual}
\end{table}

%% file: tables/variation_multilingual.tex
\begin{table}[t!]
\centering
\scriptsize
\begin{tabular}{c|c|ccc}
Loss                              & \begin{tabular}[c]{@{}c@{}}Parallel\\ Corpus\end{tabular} & Tatoeba       & BUCC          & FLORES-200   \\ \hline
\labtwo{}          & All pairs       & 0.950 & 0.979 & 0.04 \\
\labone{}         & All pairs       & \textbf{0.952} & 0.978 & 0.04 \\
\labone{} + \CrossMono{} & All pairs       & 0.948   & 0.983   & 0.04  \\ \hline
\labtwo{}          & \tten{}-\ttxx{} & 0.948 & 0.979 & 0.04 \\
\labone{}         & \tten{}-\ttxx{} & 0.942 & 0.979 & 0.05 \\
\labone{} + \CrossMono{} & \tten{}-\ttxx{} & 0.949 & \textbf{0.983} & \textbf{0.02} \\ \hline
\grayclineright{Pretrained Model} & \grayclineright{-}                                        & \grayc{0.923} & \grayc{0.981} & \grayc{0.16}
\end{tabular}
\vspace{-2mm}
\caption{Comparison of different loss on multi-lingual data, in terms of the bitext mining task performances.
}
\label{tab:ours variation multilingual}
\end{table}

%% file: tables/less_languages.tex
\begin{table}[t!]
\centering
\scriptsize
\setlength{\tabcolsep}{1.3pt}
\begin{tabular}{c|ccccc|ccc}
\multirow{2}{*}{Data} & \multicolumn{5}{c|}{STS} & \multirow{2}{*}{Tatoeba} & \multirow{2}{*}{BUCC} & \multirow{2}{*}{FLORES-200} \\ \cline{2-6}
                                 & \tten{} & \ttko{} & \ttfr{} & \ttru{} & \tten-\ttfr &       &       &      \\ \hline
\tten{}-\ttxx{}, \ttfr{}-\ttxx{} & \textbf{0.757}   & 0.694   & \textbf{0.742}   & \textbf{0.589}   & \textbf{0.765}       & \textbf{0.948} & \textbf{0.983} & \textbf{0.04} \\
\tten{}-\ttxx{}, \ttru{}-\ttxx{} & \textbf{0.757}   & \textbf{0.696}   & 0.718   & 0.586   & 0.758       & 0.946 & 0.979 & 0.07
\end{tabular}
\vspace{-2mm}
\caption{\label{tab:less language-pair}Comparison of  varying language pairs for train corpus, tested 
on bitext mining tasks and STS tasks.
The model trained on \tten{}-\ttxx{}, \ttfr{}-\ttxx{} performs better than the model trained on \tten{}-\ttxx{}, \ttru{}-\ttxx{} in most cases}
\end{table}

%% file: supplementary.tex
\appendix
\section{Full evaluation results}
\label{Appendix A: full MTEB}
\input{tables/appendix_tables}

\newpage
\section{Dataset}
\label{Appendix B.1: DATASET}

For fine-tuning the embedding model, we use translation pair from OPUS~\citep{OPUS} datasets including Tatoeba, GlobalVoices, TED2020, NewsCommentary, WikiMatrix, Europarl, OpenSubtitles2018, UNPC. We randomly sample 25k pairs from each bilingual corpus and combine them, using 20K as a train set and 5K as a validation set.
Recall that we test each model on Tatoeba dataset in MTEB~\citep{MTEB} benchmark, which is not identical to the Tatoeba dataset in OPUS. In order to make sure that the Tatoeba samples in the train and test datasets are not overlapped, 
we remove any sentences in the training data (Tatoeba dataset in OPUS) that is overlapping with the test data (Tatoeba dataset in MTEB).
The amount of training pairs contained in this filtered training dataset is given in Table~\ref{table:dataset_amount}. Our training data is available at \href{https://huggingface.co/datasets/wecover/OPUS}{https://huggingface.co/datasets/wecover/OPUS}.

\begin{table}[htb!]
\centering
\scriptsize
\begin{tabular}{c|ccccc}
pairs & \tten  & \ttko & \ttfr  & \ttja & \ttru  \\ \hline
\tten & -      & 66.8k & 153.9k & 78.5k & 133.8k \\
\ttko & 66.8k  & -     & 61.2k  & 57.6k & 58.7k  \\
\ttfr & 153.9k & 61.2k & -      & 76.8k & 132.9k \\
\ttja & 78.5k  & 57.6k & 76.8k  & -     & 76.6k  \\
\ttru & 133.8k & 58.7k & 132.9k & 76.6k & -     
\end{tabular}
\caption{Bilingual corpus statistics used in the OPUS training dataset used in our experiments. In addition, a portion of Tatoeba dataset (in OPUS) is removed, which guarantees that the test dataset does not overlap with the training dataset.}
\label{table:dataset_amount}
\end{table}

\section{Implementation details}
In this section, we describe the details of how we choose the experimental setting; especially (1) how we choose the student model and (2) how we set up multi-gpu for multi-lingual training. Under the training/evaluation processes for choosing those experimental settings, we use \tten{}-\ttko{} parallel corpus only.

\subsection{Student model selection}\label{sec:student-model-selection}

Recall that our soft contrastive learning method includes the knowledge distillation from a teacher model to a student model. We consider three student model candidates, mBERT~\citep{devlin2018bert}, XLM-R~\citep{XLM-Roberta} and $\text{mE5}_\text{base}$~\citep{wang2024me5}, and choose one that has the maximum performance, when the teacher model is fixed as $\text{mE5}_\text{base}$. Table~\ref{tab:Student} shows the performance comparison of three student models, for cases when the distillation is applied or not. ``Teacher $\rightarrow$ Student'' (\eg  ``$\text{mE5}_\text{base} \rightarrow$ mBERT'') written in Table~\ref{tab:Student} means the student model trained by the knowledge distilled from the teacher model.
One can confirm that training with our soft contrastive loss significantly increases the performance across all student models. For example, using mBERT alone only achieves the accuracy of 0.325 on Tatoeba (\tten{}-\ttko{}), which improves to 0.845 when our method is applied with  $\text{mE5}_\text{base}$ teacher model.
Since using $\text{mE5}_\text{base}$ as the student model shows the best performance, we use it as the default student model throughout the experiments shown in the main body of this paper.

\begin{table}[htb!]
\scriptsize
\setlength{\tabcolsep}{4pt}
\centering
\begin{tabular}{c|c|c|cc}
Model                                                                 & \begin{tabular}[c]{@{}c@{}}Model\\ Size\end{tabular} & \begin{tabular}[c]{@{}c@{}}Tatoeba\\ (\tten-\ttko)\end{tabular} & \begin{tabular}[c]{@{}c@{}}STS\\ (\tten)\end{tabular} & \begin{tabular}[c]{@{}c@{}}STS\\ (\ttko)\end{tabular} \\ \hline
mBERT                                                                 & 108M                                                 & 0.325                                                           & 0.519                                                 & 0.546                                                 \\
$\text{mE5}_\text{base}$ (\labtwo{}) $\rightarrow$ mBERT                    & 108M                                                 & 0.845                                                           & 0.635                                                 & 0.640                                                 \\ \hline
XLM-R                                                                 & 270M                                                 & 0.220                                                           & 0.429                                                 & 0.511                                                 \\
$\text{mE5}_\text{base}$ (\labtwo{}) $\rightarrow$ XLM-R                    & 270M                                                 & 0.904                                                           & 0.698                                                 & 0.707                                                 \\ \hline
$\text{mE5}_\text{base}$                                              & 270M                                                 & 0.875                                                           & 0.802                                                 & 0.777                                                 \\
$\text{mE5}_\text{base}$ (\labtwo{}) $\rightarrow$ $\text{mE5}_\text{base}$ & 270M                                                 & \textbf{0.912}                                                  & \textbf{0.771}                                        & \textbf{0.757}                                       
\end{tabular}
\caption{Performance on bitext-retrieval tasks, tested on various student models, when the teacher model is fixed to $\text{mE5}_\text{base}$. This result shows that using $\text{mE5}_\text{base}$ as the student model performs the best, which guides us to use it as default.}
\label{tab:Student}
\end{table}

\subsection{Multi-GPU setting}
\label{Appendix B.4: CA}
The cross-accelerator method is considered in the contrastive learning of LaBSE~\citep{LABSE}, which enables enlarging both the negative sample size and the global batch size, thus improving the performance of contrastive learning. We test the effect of using cross-accelerator (instead of in-batch negative sampling) in Table~\ref{tab:multi-gpu}. One can confirm that the cross-accelerator setting outperforms in-batch setting for all benchmarks.
Here, we implemented cross-accelerator based on rocketQA
\footnote{\url{https://github.com/PaddlePaddle/RocketQA.}}
~\citep{qu2020rocketqa} approach. 
\begin{table}[t!]
\centering
\scriptsize
\begin{tabular}{c|ccc}
                  & \begin{tabular}[c]{@{}c@{}}Tatoeba\\ (\texttt{en-ko})\end{tabular} & \begin{tabular}[c]{@{}c@{}}STS\\ (\tten)\end{tabular} & \begin{tabular}[c]{@{}c@{}}STS\\ (\ttko)\end{tabular} \\ \hline
in-batch          & 0.887                                                     & 0.695                                              & 0.699                                              \\
cross-accelerator & \textbf{0.904}                                            & \textbf{0.698}                                     & \textbf{0.707}                                    
\end{tabular}
\caption{MTEB performances of the in-batch negative sampling method and the cross-accelerator approach.}
\label{tab:multi-gpu}
\end{table}

%% file: tables/appendix_tables.tex
\begin{table}[htb!]
\centering
\tiny
\setlength{\tabcolsep}{2pt}
\begin{tabular}{c|c|ccccc}
\multirow{2}{*}{Model} &
  \multirow{2}{*}{\begin{tabular}[c]{@{}c@{}}Parallel\\ Corpus\end{tabular}} &
  \multicolumn{5}{c}{Tatoeba} \\ \cline{3-7} 
 &
   &
  \multicolumn{1}{c|}{avg} &
  \texttt{en-ko} &
  \texttt{en-fr} &
  \texttt{en-ja} &
  \texttt{en-ru} \\ \hline
$\text{mE5}_\text{base}$ &
  - &
  \multicolumn{1}{c|}{0.923} &
  0.875 &
  0.951 &
  0.931 &
  0.936 \\
MPNet-multilingual &
  - &
  \multicolumn{1}{c|}{0.945} &
  \textbf{0.947} &
  0.946 &
  0.941 &
  0.945 \\
LASER2 &
  - &
  \multicolumn{1}{c|}{0.939} &
  0.906 &
  0.955 &
  0.952 &
  0.942 \\
LaBSE &
  - &
  \multicolumn{1}{c|}{0.948} &
  0.926 &
  0.954 &
  0.960 &
  \textbf{0.952} \\ \hline
$\text{mE5}_\text{base}$ + Ours (\labone{})          & All pairs      & \multicolumn{1}{c|}{\textbf{0.952}} & 0.922 & \textbf{0.963} & \textbf{0.972} & 0.950 \\
$\text{mE5}_\text{base}$ + Ours (\labtwo{}) &
  All pairs &
  \multicolumn{1}{c|}{{{\ul 0.950}}} &
  0.923 &
  0.959 &
  {{\ul 0.966}} &
  \textbf{0.952} \\
$\text{mE5}_\text{base}$ + Ours (\labone{}) + \CrossMono{}          & All pairs      & \multicolumn{1}{c|}{0.948} & 0.921 & 0.957 & 0.964 & 0.950 \\
$\text{mE5}_\text{base}$ + Ours (\labone{}) &
  \texttt{en-xx}, \texttt{fr-xx} &
  \multicolumn{1}{c|}{0.948} &
  0.921 &
  0.958 &
  0.961 &
  0.950 \\
$\text{mE5}_\text{base}$ + Ours (\labone{}) &
  \texttt{en-xx}, \texttt{ru-xx} &
  \multicolumn{1}{c|}{0.946} &
  0.910 &
  {{\ul 0.961}} &
  0.965 &
  0.948 \\
$\text{mE5}_\text{base}$ + Ours (\labone{}) &
  \texttt{en-xx} &
  \multicolumn{1}{c|}{0.942} &
  0.912 &
  0.955 &
  0.955 &
  0.945 \\
$\text{mE5}_\text{base}$ + Ours (\labtwo{}) &
  \texttt{en-xx} &
  \multicolumn{1}{c|}{0.948} &
  0.927 &
  0.958 &
  0.957 &
  0.948 \\
$\text{mE5}_\text{base}$ + Ours (\labone{}) + \CrossMono{} & \texttt{en-xx} & \multicolumn{1}{c|}{0.949}          & {{\ul 0.929}} & \textbf{0.963} & 0.951          & {{\ul 0.951}} \\
$\text{mE5}_\text{base}$ + MSE &
  \texttt{en-xx} &
  \multicolumn{1}{c|}{0.942} &
  0.916 &
  0.956 &
  0.947 &
  0.948
\end{tabular}
\vspace{-2mm}
\caption{Accuracy (↑) measured on Tatoeba dataset, for all tested pairs}
\label{tab:multilingual tatoeba full result}
\end{table}

\begin{table}[htb!]
\centering
\scriptsize
\setlength{\tabcolsep}{2.8pt}
\begin{tabular}{c|c|ccc}
\multirow{2}{*}{Model} &
  \multirow{2}{*}{\begin{tabular}[c]{@{}c@{}}Parallel\\ Corpus\end{tabular}} &
  \multicolumn{3}{c}{BUCC} \\ \cline{3-5} 
                                      &                & \multicolumn{1}{c|}{avg}            & \texttt{en-fr} & \texttt{en-ru} \\ \hline
$\text{mE5}_\text{base}$              & -              & \multicolumn{1}{c|}{0.981}          & 0.984          & 0.978          \\
MPNet-multilingual                    & -              & \multicolumn{1}{c|}{0.970}          & 0.972          & 0.968          \\
LASER2                                & -              & \multicolumn{1}{c|}{0.981}          & 0.985          & 0.977          \\
LaBSE                                 & -              & \multicolumn{1}{c|}{\textbf{0.985}} & \textbf{0.991} & \textbf{0.980} \\ \hline
$\text{mE5}_\text{base}$ + Ours (\labone{})   & All pairs      & \multicolumn{1}{c|}{0.977}          & 0.979          & 0.976          \\
$\text{mE5}_\text{base}$ + Ours (\labtwo{}) & All pairs      & \multicolumn{1}{c|}{0.979}          & 0.982          & 0.976          \\
$\text{mE5}_\text{base}$ + Ours (\labone{}) + \CrossMono{}   & All pairs      & \multicolumn{1}{c|}{{\ul 0.983}}          & {\ul 0.987}          & {\ul 0.979}          \\
$\text{mE5}_\text{base}$ + Ours (\labone{}) &
  \texttt{en-xx}, \texttt{fr-xx} &
  \multicolumn{1}{c|}{{\ul 0.983}} &
  {\ul 0.987} &
  {\ul 0.979} \\
$\text{mE5}_\text{base}$ + Ours (\labone{}) &
  \texttt{en-xx}, \texttt{ru-xx} &
  \multicolumn{1}{c|}{0.979} &
  0.981 &
  0.977 \\
$\text{mE5}_\text{base}$ + Ours (\labone{})   & \texttt{en-xx} & \multicolumn{1}{c|}{0.979}          & 0.980          & 0.977          \\
$\text{mE5}_\text{base}$ + Ours (\labtwo{}) &
  \texttt{en-xx} &
  \multicolumn{1}{c|}{0.979} &
  0.982 &
  0.977 \\
$\text{mE5}_\text{base}$ + Ours (\labone{}) + \CrossMono{} &
  \texttt{en-xx} &
  \multicolumn{1}{c|}{{\ul 0.983}} &
  0.986 &
  {\ul 0.979} \\
$\text{mE5}_\text{base}$ + MSE        & \texttt{en-xx} & \multicolumn{1}{c|}{0.975}          & 0.974          & 0.976         
\end{tabular}
\vspace{-2mm}
\caption{Accuracy (↑) measured on BUCC dataset, for all tested pairs.}
\label{tab:multilingual BUCC full result}
\end{table}

\begin{table}[htb!]
\centering
\tiny
\setlength{\tabcolsep}{2pt}
\begin{tabular}{c|c|cccccc}
\multirow{2}{*}{Model} & \multirow{2}{*}{\begin{tabular}[c]{@{}c@{}}Parallel\\ Corpus\end{tabular}} & \multicolumn{6}{c}{FLORES-200} \\ \cline{3-8} 
                                       &                                & \multicolumn{1}{c|}{avg}           & \tten & \ttko & \ttfr & \ttja & \ttru \\ \hline
$\text{mE5}_\text{base}$               & -                              & \multicolumn{1}{c|}{0.16}          & 0.05  & 0.40  & 0.12  & 0.02  & 0.20  \\
MPNet-multilingual                     & -                              & \multicolumn{1}{c|}{0.28}          & 0.17  & 0.49  & 0.35  & 0.17  & 0.20  \\
LASER2                                 & -                              & \multicolumn{1}{c|}{0.20}          & 0.10  & 0.32  & 0.10  & 0.10  & 0.40  \\
LaBSE                                  & -                              & \multicolumn{1}{c|}{\textbf{0.01}} & {\textbf{0.00}}  & {\textbf{0.00}}  & 0.02  & {\textbf{0.00}}  & {\textbf{0.02}}  \\ \hline
$\text{mE5}_\text{base}$ + Ours (\labone{})    & All pairs                      & \multicolumn{1}{c|}{0.04}          & 0.02  & 0.05  & 0.07  & 0.05  & {\textbf{0.02}}  \\
$\text{mE5}_\text{base}$ + Ours (\labtwo{})  & All pairs                      & \multicolumn{1}{c|}{0.04}          & 0.02  & 0.05  & 0.05  & 0.05  & {\textbf{0.02}}  \\
$\text{mE5}_\text{base}$ + Ours (\labone{}) + \CrossMono{}    & All pairs                      & \multicolumn{1}{c|}{0.04}          & 0.02  & 0.02  & {\textbf{0.00}}  & 0.15  & {\textbf{0.02}}  \\
$\text{mE5}_\text{base}$ + Ours (\labone{})    & \texttt{en-xx}, \texttt{fr-xx} & \multicolumn{1}{c|}{0.04}          & 0.02  & 0.12  & 0.02  & {\textbf{0.00}}  & 0.05  \\
$\text{mE5}_\text{base}$ + Ours (\labone{})    & \texttt{en-xx}, \texttt{ru-xx} & \multicolumn{1}{c|}{0.07}          & 0.02  & 0.12  & 0.10  & 0.02  & {\textbf{0.02}}  \\
$\text{mE5}_\text{base}$ + Ours (\labone{})    & \texttt{en-xx}                 & \multicolumn{1}{c|}{0.05}          & {\textbf{0.00}}  & 0.12  & 0.02  & 0.02  & 0.07  \\
$\text{mE5}_\text{base}$ + Ours (\labtwo{})    & \texttt{en-xx}                 & \multicolumn{1}{c|}{0.04}          & 0.02  & 0.05  & 0.02  & 0.10  & {\textbf{0.02}}  \\
$\text{mE5}_\text{base}$ + Ours (\labone{}) + \CrossMono{} & \texttt{en-xx}                 & \multicolumn{1}{c|} {0.02}    & {\textbf{0.00}}  & 0.02  & 0.02  & 0.02  & {\textbf{0.02}}  \\
$\text{mE5}_\text{base}$ + MSE         & \texttt{en-xx}                 & \multicolumn{1}{c|}{0.05}          & 0.02  & 0.12  & 0.05  & 0.02  & 0.05 
\end{tabular}
\vspace{-2mm}
\caption{xSIM error rate (↓) measured on FLORES-200
dataset, for all tested languages}
\label{tab:multilingual xsim full result}
\end{table}

\begin{table}[h]
\centering
\tiny
\setlength{\tabcolsep}{2pt}
\begin{tabular}{c|c|ccccc}
\multirow{2}{*}{Model} & \multirow{2}{*}{\begin{tabular}[c]{@{}c@{}}Parallel\\ Corpus\end{tabular}} & \multicolumn{5}{c}{STS} \\ \cline{3-7} 
                                             &                                & \tten & \ttko & \ttfr & \ttru & \texttt{en-fr} \\ \hline
$\text{mE5}_\text{base}$                     & -                              & {\textbf{0.802}} & 0.777 & {\textbf{0.781}} & {{\ul 0.615}} & {{\ul 0.813}}          \\
MPNet-multilingual                           & -                              & {\textbf{0.802}} & {\textbf{0.834}} & 0.542 & 0.454 & 0.812          \\
LASER2                                       & -                              & 0.635 & 0.705 & 0.586 & 0.392 & 0.708          \\
LaBSE                                        & -                              & 0.694 & 0.753 & 0.707 & 0.573 & 0.751          \\ \hline
$\text{mE5}_\text{base}$ + Ours (\labone{})          & All pairs                      & 0.741 & 0.688 & 0.695 & 0.580 & 0.761          \\
$\text{mE5}_\text{base}$ + Ours (\labtwo{})        & All pairs                      & 0.738 & 0.707 & 0.691 & 0.557 & 0.751          \\
$\text{mE5}_\text{base}$ + Ours (\labone{}) + \CrossMono{}         & All pairs                      & 0.774 & 0.770 & 0.695 & 0.595 & 0.775          \\
$\text{mE5}_\text{base}$ + Ours (\labone{})          & \texttt{en-xx}, \texttt{fr-xx} & 0.757 & 0.694 & 0.742 & 0.589 & 0.765          \\
$\text{mE5}_\text{base}$ + Ours (\labone{})          & \texttt{en-xx}, \texttt{ru-xx} & 0.757 & 0.696 & 0.718 & 0.586 & 0.758          \\
$\text{mE5}_\text{base}$ + Ours (\labone{})          & \texttt{en-xx}                 & 0.764 & 0.726 & 0.728 & 0.589 & 0.779          \\
$\text{mE5}_\text{base}$ + Ours (\labtwo{})          & \texttt{en-xx}                 & 0.763 & 0.742 & {{\ul 0.760}} & 0.596 & 0.782          \\
$\text{mE5}_\text{base}$ + Ours (\labone{}) + \CrossMono{} & \texttt{en-xx}                 & 0.782 & 0.769 & {{\ul 0.760}} & {\textbf{0.624}} & 0.798          \\
$\text{mE5}_\text{base}$ + MSE               & \texttt{en-xx}                 & {{\ul 0.800}} & {{\ul 0.790}} & 0.731 & 0.601 & {\textbf{0.816}}         
\end{tabular}
\vspace{-2mm}
\caption{\label{tab:multilingual STS full result}STS results (↑) measured on different languages.}
\end{table}